\documentclass[runningheads]{llncs}
\usepackage[T1]{fontenc}
\usepackage{graphicx}
\usepackage{booktabs}
\usepackage{adjustbox}
\usepackage{hyperref}
\usepackage{color}
\usepackage{xcolor}
\usepackage{makecell}
\usepackage{tabularx}

\usepackage[acronym,nonumberlist]{glossaries}
\glsdisablehyper
\newacronym{cnn}{CNN}{Convolutional Neural Network}
\newacronym{eo}{EO}{electro-optical}
\newacronym{fm}{FM}{foundation model}
\newacronym{genai}{GenAI}{generative artificial intelligence}
\newacronym{gis}{GIS}{geographic information system}
\newacronym{gpu}{GPU}{graphics processing unit}
\newacronym{gsd}{GSD}{ground sampling distance}
\newacronym{g-daug}{G-DAUG}{Geospatial Data Augmentation}
\newacronym{lr}{LR}{learning rate}
\newacronym{mae}{MAE}{masked autoencoder}
\newacronym{mse}{MSE}{mean squared error}
\newacronym{ml}{ML}{machine learning}
\newacronym{msi}{MSI}{multispectral imagery}
\newacronym{mtp}{MTP}{multi-task pretraining}
\newacronym{nlp}{NLP}{natural language processing}
\newacronym{osm}{OSM}{OpenStreetMap}
\newacronym{pan}{PAN}{panchromatic}
\newacronym{rs}{RS}{remote sensing}
\newacronym{rvsa}{RVSA}{rotated varied-size attention}
\newacronym{sar}{SAR}{synthetic aperture radar}
\newacronym{scalemae}{ScaleMAE}{scale-aware masked autoencoder}
\newacronym{vit}{ViT}{vision transformer}
\newacronym{wsd}{WSD}{Warmup, Stable, Decay}
\begin{document}
\title{Scaling Remote Sensing Foundation Models: Data Domain Tradeoffs at the Peta-Scale}
\titlerunning{Scaling Remote Sensing Foundation Models}
%s
\author{
Charith Wickrema\inst{1}\orcidID{0009-0006-2679-4453} \and
Eliza Mace\inst{1}\orcidID{0009-0000-0305-7449} \and
Hunter Brown\inst{1}\orcidID{0009-0005-8013-1207} \and
Heidys Cabrera\inst{1}\orcidID{0009-0000-9811-8856} \and
Nick Krall\inst{1}\orcidID{0009-0007-1001-2071}\and
Matthew O'Neill\inst{1}\orcidID{0009-0000-8876-6956} \and
Shivangi Sarkar\inst{1}\orcidID{0009-0003-7281-7511} \and
Lowell Weissman\inst{1}\orcidID{0009-0006-5253-9843} \and
Eric Hughes\inst{1}\orcidID{0009-0002-1509-353X} \and
Guido Zarrella\inst{1}\orcidID{0000-0002-5770-0257} }
\authorrunning{C. Wickrema et al.}
\institute{The MITRE Corporation, McLean VA 22102, USA\\
\email{\{cwickrema,emace,hbrown,hcabrera,nkrall,moneill,\\
ssarkar,lweissman,hughes,jzarrella\}@mitre.org}\\
\url{https://www.mitre.org/}}
\maketitle              % typeset the header of the contribution
\begin{abstract}
    We explore the scaling behaviors of artificial intelligence to establish practical techniques for training foundation models on high-resolution \gls{eo} datasets that exceed the current state-of-the-art scale by orders of magnitude.
    Modern multimodal \gls{ml} applications, such as \gls{genai} systems for image captioning, search, and reasoning, depend on robust, domain-specialized encoders for non-text modalities.
    In natural image domains where internet-scale data is plentiful, well established scaling laws help optimize the joint scaling of model capacity, training compute, and dataset size.
    Unfortunately these relationships are much less well understood in high-value domains like \gls{rs}.
    Using over a quadrillion pixels of commercial satellite \gls{eo} data and MITRE’s Federal AI Sandbox, we train progressively larger \gls{vit} backbones, report successes and failure modes observed at petascale, and analyze implications for bridging domain gaps across additional \gls{rs} modalities. 
    We observe that even at this scale, performance is consistent with a data-limited regime rather than a model parameter-limited one. 
    These practical insights are intended to inform data collection strategies, compute budgets, and optimization schedules that advance the future development of frontier scale \gls{rs} foundation models.
\end{abstract}
%
%\glsresetall
%
%
\section{Introduction}
\label{section:intro}
    Recent advances in \gls{ml} have been propelled by the development of large-scale models and foundation architectures, particularly in the context of \gls{genai} and multimodal applications. 
    Central to the success of these systems are performant data encoders which transform raw data into representations useful in downstream tasks. 
    While scaling laws and architectural best practices for \glspl{vit} have been extensively studied in domains such as natural imagery \cite{scalingvits} \cite{nlpscaling}, their applicability to specialized domains like \gls{rs} remains underexplored.
    
    Remote sensing data, especially from \gls{eo} commercial satellites, presents unique challenges and opportunities for \gls{ml}. 
    Unlike natural images, \gls{rs} data is characterized by diverse spectral, spatial, and temporal properties, as well as limited availability of labeled, in-domain training samples. 
    This scarcity has hindered the development and scaling of \glspl{fm} tailored to \gls{rs}, leaving several fundamental questions unanswered. 
    Notably, the relationship between training dataset size and model performance at scale for in-domain \gls{rs} \glspl{fm} is poorly understood.
    
    Addressing these gaps is critical for advancing the state of the art in \gls{rs} \gls{ml} applications, which underpin a wide range of societal and scientific needs, from environmental monitoring to urban planning. 
    To this end, we leverage an unprecedented scale of \gls{eo} satellite imagery totaling over a peta-pixel and employ \gls{mtp} \cite{mtp} and \gls{scalemae} \cite{scalemae} foundation model architectures.
    By integrating supervised optimization with open-source \gls{gis} derived labels, we are able to construct large, high-quality training datasets. 
    Furthermore, by utilizing large-scale \glspl{gpu} resources, specifically an NVIDIA DGX H100 SuperPOD in MITRE's Federal AI Sandbox, we systematically scale \gls{vit} training regimes to assess whether modern frontier \gls{rs} foundation models remain constrained by data availability.
    
    In this work, we empirically characterize the scaling behaviors of \gls{vit}-based \gls{rs} \glspl{fm}, evaluating their performance and analyzing their capacity to address domain gaps across additional geospatial modalities. 
    Our findings provide new insights into the design and training of remote sensing \glspl{fm}, with implications for both the \gls{ml} and \gls{rs} communities.
    
    \subsection{Guiding Questions}
        This study is organized around four guiding questions central to scaling \gls{rs} encoders: (a) under what conditions increasing batch size improves training time or convergence; (b) which \gls{lr} schedules remain stable for continual pretraining of \gls{rs} models at scale; (c) how performance scales with in-domain dataset size and the rate at which gains taper; and (d) how predictably performance improves when scaling \gls{vit} model parameters. These questions structure our experimental design and analyses to develop practical strategies for future \gls{rs} \glspl{fm} training.

    \subsection{Motivation}
    \label{section:motivation}
        The unprecedented scale and diversity of modern \gls{rs} datasets enable the training of large, domain-specific models that have the potential to outperform generic, out-of-domain encoders on \gls{rs} tasks. 
        However, the unique characteristics of \gls{rs} data such as its spectral complexity, geographic diversity, and limited labeled samples raise important questions about the optimal strategies for model scaling in this context.
        
        A key open question is whether current frontier-scale \gls{rs} models adequately saturate performance, in other words, whether continued scaling offers opportunity for improvement or if diminishing returns emerge as model and dataset sizes increase beyond current state-of-the-art implementations.
        Unlike domains such as \gls{nlp} or natural image understanding, where increasing model size unlocks richer context-aware representations and better captures longer-range dependencies--- due in part to the inherent complexity and compositionality in the data---\gls{rs} imagery may present intrinsic limitations. 
        The set of features that can be learned from overhead imagery could be fundamentally constrained by the spatial resolution and the finite variety of observable objects and patterns within a given scene. 
        As a result, there may be an upper bound to the representational capacity required for effective \gls{rs} analysis, beyond which further scaling yields marginal gains. 
        Furthermore, while larger models in \gls{nlp} often benefit from expanded context windows that enable reasoning over longer sequences, the spatial context in overhead imagery is typically limited by the maximum image size available from traditional sensors. 
        This raises the possibility that simply increasing model size may not indefinitely improve performance in \gls{rs} tasks, as the available contextual information does not scale in the same manner as in other domains.
        
        Addressing these questions is essential for maximizing the impact of foundation models in remote sensing. 
        By systematically exploring the interplay between model size, training data composition, and model performance, we aim to provide actionable insights for both the \gls{ml} and \gls{rs} communities, informing the next generation of scalable, high-performance geospatially-aware models.
        Given these constraints, we treat scaling as an engineering process: identify the levers that matter (data composition, optimizer dynamics, throughput), stress-test them at new scales, and document the regimes where progress is linear versus where it stalls or destabilizes. 
        The remainder of the paper follows this process, moving from related literature to our dataset design, controlled experiments, and the lessons derived from their outcomes.
    
    \subsection{Contributions}
        We contribute a process-oriented account of scaling \gls{rs} foundation models:
        \begin{itemize}
            \item Lessons for throughput scaling: early- to mid-stage training is batch-insensitive beyond moderate thresholds; prioritize data diversity and \gls{lr} scheduling over aggressive batch increases.
            \item Stable optimization schedules: a \gls{wsd} regime with moderate base learning rate (LR) provides robust convergence; we document divergence behaviors at higher \glspl{lr} and propose experiments to avoid wasting compute on unstable runs.
            \item Empirical data-scaling law: within 5k–1M full scene \gls{rs} images, loss follows a consistent power-law with diminishing but non-saturated returns, supporting larger compute- and data-planning for \gls{rs} pretraining.
            \item Empirical model scaling laws: across 86M-1.8B parameters we see effectively flat returns indicating that model scaling alone provides negligible benefit under typical data size regimes.
            \item Scalable weak supervision: a reproducible \gls{g-daug} pipeline that pairs image chips with automatic enrichments allows exploration of supervision techniques relevant at global scale.
        \end{itemize}

\section{Related Work}
\label{section:related}
    \subsection{Foundation Models and Scaling Laws in Vision}
        The adoption of transformer units \cite{attention} \cite{vits} revolutionized representation learning by introducing self-attention mechanisms that scale efficiently with data and compute. 
        This paradigm shift enabled the rise of foundation models, large, pretrained systems capable of generalizing across diverse downstream tasks with minimal adaptation. 
        Studies on neural network scaling later showed that model performance frequently follows power-law relationships with respect to model size, dataset size, and compute. 
        Previous research \cite{nlpscaling} demonstrated these trends for transformer-based language models showing that loss scales predictably and that compute can be optimally distributed across parameters, data, and training steps.
        In vision, research \cite{scalingvits} extended scaling analysis to \glspl{vit}, finding a compute performance pareto frontier and double-saturating power law indicating that data remains the limiting factor even for billion-scale image datasets. 
        Larger models continue to benefit from more data, while smaller ones quickly saturate. 
        Scaling-law methods were quickly refined through improved curve fitting \cite{revistneuralscaling}, enhancing extrapolation across new scales and modalities and enabling more accurate prediction of compute efficiency. 
        Together these works established scaling laws as a predictive framework for large-scale model design. 
        Our study extends this foundation to geospatial foundation models, where domain heterogeneity, sensor diversity, and spatial temporal structure may yield distinct scaling dynamics. 

    \subsection{Vision for Remote Sensing}
        Trends towards foundational modeling in vision and language are also reflected in the research community's move towards geospatial foundation models for remote sensing. 
        These reflect a rapid shift from narrow, task-specific networks to large-scale, general-purpose models trained on globally distributed satellite imagery. 
        Early \gls{rs} models were limited by modest dataset size, handcrafted features, and single-task training, but the rise of transformers and large-scale self-supervised learning has enabled development of unified models capable of transferring across sensors, spatial scales, and geographies. 
        
        To address the diversity and scale of earth observation data, researchers have begun to assemble increasingly larger, more heterogenous datasets and corresponding pretraining strategies. 
        SatlasPretrain \cite{satlaspretrain} created one of the largest labeled \gls{rs} pretraining datasets to date with multiple sensors and tasks: $1\times 10^{13}$ pixels over 21 million $\mathrm{km}^2$ with 137 distinct categories. 
        The model demonstrated that large-scale joint pretraining improves downstream performance on a variety of segmentation, object detection, and change detection tasks marking an early step toward foundation-scale generalization in \gls{rs}. 
        Similarly, Prithvi \cite{prithvi} explored self-supervised pretraining for multispectral imagery at continental scale (4.2M globally distributed samples), revealing that transformer models can learn general geospatial representations transferable across multiple datasets. 
        
        Architecturally, recent work has been focused on incorporating scale awareness and geospatial priors into vision transformers. 
         \gls{scalemae} \cite{scalemae} introduced a scale-aware masked autoencoder that anchors positional encoding to real-world ground sample distance (GSD), enabling consistent multiresolution learning across diverse imagery sources. 
        Other work proposed Multi-Task Pretraining (MTP) \cite{mtp} frameworks that unify segmentation, detection, and classification objectives, while SkySense \cite{skysense} extended foundational pretraining to billions of parameters and multimodal inputs, both temporal and contextual, reflecting the domains' upward progress. 
        Efforts that have explicitly expanded parameter scaling, training transformer backbones ranging from 86M to 2B parameters \cite{billionscale}, observed consistent performance in object detection and segmentation regardless of backbone size, likely due to a lack of commensurate data scaling with parameter scaling, as they leveraged the same 1M examples as the original \gls{mtp} effort \cite{mtp}. 
        Meanwhile, DINOv3 \cite{dinov3}, provides evidence that scaling model and data size in self-supervision can produce dense transferable representations that generalize to \gls{rs} domains, starting both from natural and \gls{rs} images, with corpuses of 1.7B image chips and 493M image chips, respectively.
        The increase in \gls{rs} dataset sizes across the field is summarized in table \ref{tab:dataset_comparisons}.

    \subsection{Summary and Gaps}
        Overall, these developments highlight a clear trend: remote sensing has entered a new scale of vision FM development. 
        With increasing dataset size, parameter counts, and compute budgets, our next step is to quantitatively characterize scaling laws to determine how performance scales within the unique characteristics of geospatial imagery. 
        Following the success of DINOv3 \cite{dinov3} in creating relevant \gls{rs} representations, the field would benefit from precise quantification of training tradeoffs to allow for right-sizing of datasets, model architectures, and compute budgets for specific domain needs.
        Most prior work emphasizes best-case performance and model design, while comparatively little documents the operational regimes where scaling stalls, destabilizes, or provides marginal returns in \gls{rs}. 
        Our study complements these advances by quantifying when and how scaling pays off in \gls{rs}, and by reporting the failure modes and lessons learned that turn scaling laws into practical training practice.

\section{Dataset}
    \label{section:data}
    We constructed Akupara, a global remote sensing dataset comprising over a peta-pixel ($1\times10^{15}$) of high-resolution Maxar satellite imagery, the largest remote sensing dataset to date for FM training. 
    The ``Akupara-1M'' collection contains over 1 million full satellite images (equivalent to $\sim4$ billion $512\times512$ chips) with 415 million $\mathrm{km}^2$ of gross coverage (excluding USA due to collection constraints), with both single-channel panchromatic (PAN) imagery and eight-band multi-spectral imagery (MSI), and a ground sample distance (\gls{gsd}) range of 0.3 meters to 1.2 meters.
    Images were collected primarily from from 2020-2024 with <2\% of imagery from 2007-2020.
    Table \ref{tab:dataset_comparisons} shows a comparison of Akupara to other large remote sensing datasets.
    \begin{figure}[h!]
        \centering
        \includegraphics[width=1.0\linewidth]{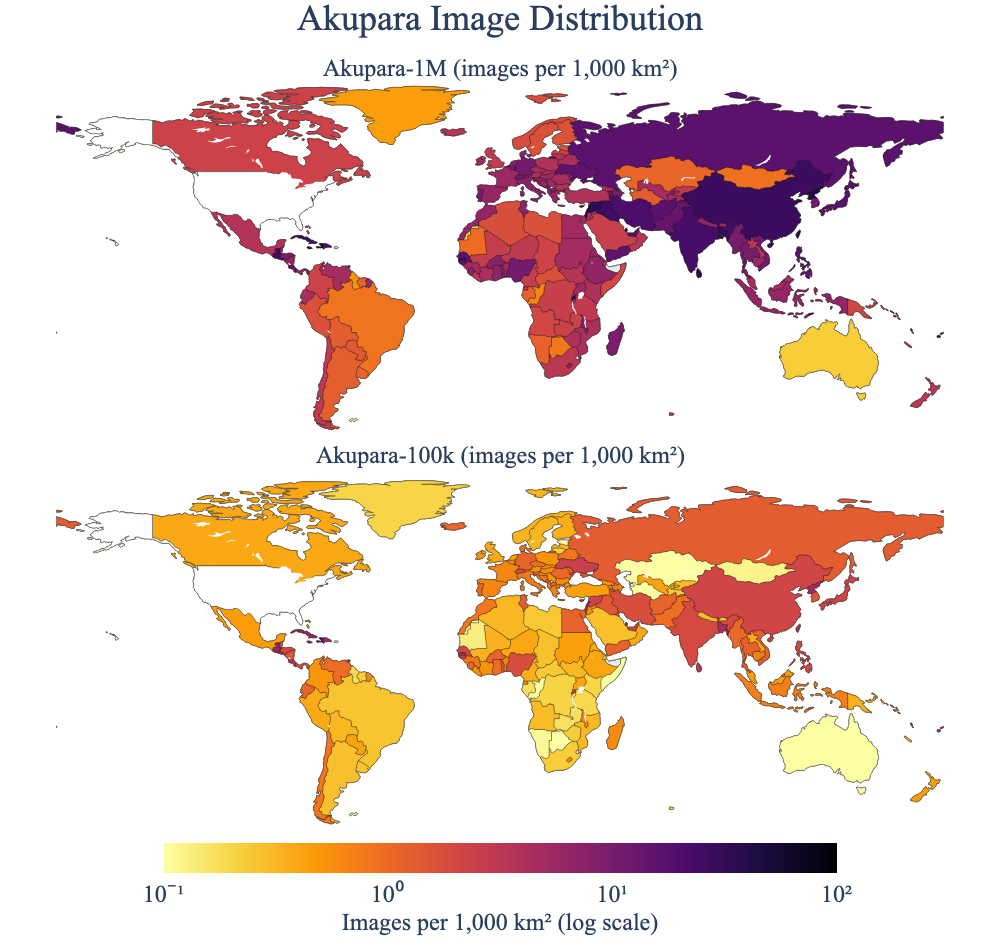}
        \caption{Visualization of the number of images per 1,000 $\mathrm{km}^2$ in both Akupara-1M and Akupara-100k. The Akupara-100k dataset is subsampled from the Akupara-1M dataset based on maximizing diversity of sensor and land area attributes. Although not visualized, collections over bodies of water, coastlines, islands, and Antarctica were also included in the dataset.}
        \label{fig:PetaSatDataDistributions}
    \end{figure}
    
    To create representative working subsets, like Akupara-(100k, 50k, 5k, etc.), for analysis and model training, we employed a stratified sampling framework combining population weighting and attribute balancing, an example of which is shown in figure \ref{fig:PetaSatDataDistributions}.
    Population density values were sampled at each image center point, and both numeric (e.g. obliquity angle, elevation variance, population density) and categorical (e.g., sensor type, time of day, land-cover class) attributes were normalized into quantile-based bins.
    Iterative proportional fitting (ranking) was used to calibrate sampling weights toward uniform marginal distributions across attributes.
    Using these values, a weighted, non-replacement sampling procedure generated fixed-size subsets with near-uniform coverage across geographic, temporal, and environmental domains. 
    Because performance in \gls{rs} appears data-limited yet feature-bounded, we opted to perform stratified sampling targeting geographic, temporal, and environmental balance so that `more data' expands the domain of variation rather than duplicating narrow regimes.
    This design choice directly supports our scaling-law analysis, ensuring that improvements with scale reflect expanded coverage rather than repeated content.
    \begin{table}[h!]
        \begin{tabularx}{\textwidth}{
        l
        >{\centering\arraybackslash}X 
        >{\centering\arraybackslash}X 
        >{\centering\arraybackslash}p{2cm}
        >{\centering\arraybackslash}X 
        >{\centering\arraybackslash}X
        }
            \toprule
            \thead{Dataset} & \thead{Num.\\pixels} &\thead{Num.\\classes} & \thead{Num.\\labels} & \thead{GSD\\(m)} & \thead{Land area\\($\mathrm{km}^2$)}\\
            \midrule
            BigEarthNet & 1e9  & 43 & 2,000 & 10-60 & 850,000 \\
            Million-AID & 4e9  & 51 & 37,000 & 0.5-153 & 18,000 \\
            FMoW & 4e11 & 63 & 417,000 & 0.5 & 1,748,000 \\
            SatlasPretrain & 1e13 & 137 & 302,222,000 & 0.5-60 & 21,320,000 \\
            DINOv3 Sat & 1e14  & - & - & 0.6 & 46,000 \\
            Akupara-100k (ours) & 7e13  & 37 & 3,800,000,000 & 0.3-1.2 & 32,677,000 \\
            Akupara-1M (ours) & 1e15  & 37 & 49,390,000,000* & 0.3-1.2 & 415,951,000 \\
            \bottomrule
        \end{tabularx}
        \caption{Akupara-1M is larger in pixel volume and land area than existing remote sensing datasets, and at a higher resolution (i.e., lower  \gls{gsd}). Labels are number of unique labels, and $\mathrm{km}^2$ is area covered. *Akupara-1M has estimated number of labels as processing is ongoing within compute resource constraints.}
        \label{tab:dataset_comparisons}
    \end{table}
    
    \subsection{Label Generation}
\begin{figure}[h!]
            \centering
            \includegraphics[width=1.0\linewidth]{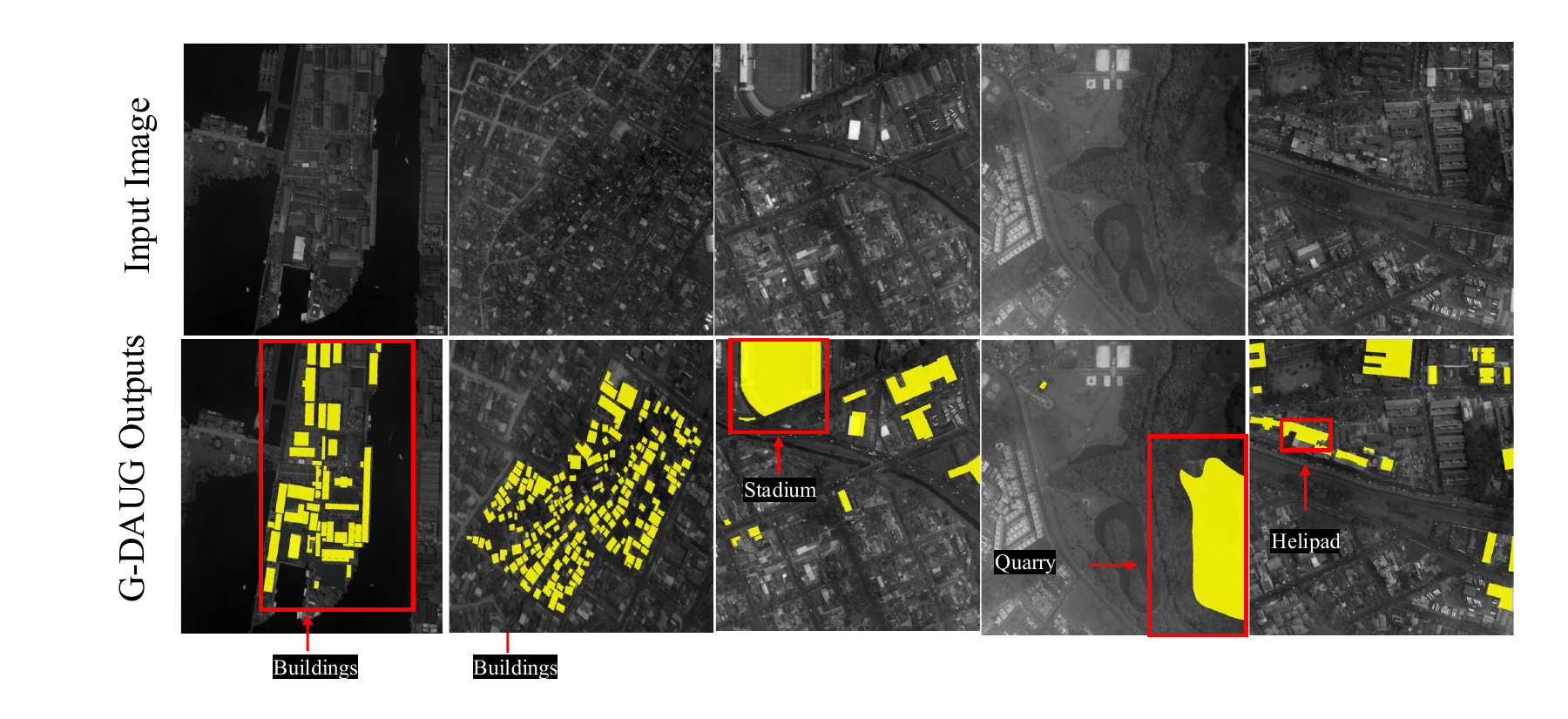}
            \caption{The \gls{g-daug} pipeline leverages \gls{osm} data and segmentation algorithms to automatically create geospatial data labels. We create pixelwise segmentation masks, as shown, which are also converted bounding boxes and instance segmentation masks for additional supervisory tasks.}
            \label{fig:GDAUG_Ex}
          \end{figure}

        We developed the automated \gls{g-daug} pipeline, a scalable geospatial labeling framework based on the SatlasPretrain \cite{satlaspretrain} labeling method for large-scale weak supervision and automated annotation.
        Each $1024\times1024$ image chip is paired with \gls{osm} vector data retrieved via PostGIS queries within the chip's geographic footprint.
        The pipeline converts chip pixel coordinates to geographic bounds using metadata from the source orthorectified imagery, then filters \gls{osm} geometries by a curated ontology of 37 object classes (based on empirical findings in \cite{satlaspretrain} that reveal categories with higher fidelity labels).
        Geometries are reprojected to the image coordinate reference system, clipped to the chip extent, and then rasterized into high-resolution binary segmentation masks. 
        We produce instance segmentation masks, semantic segmentation masks, and rotated bounding boxes labels (as shown in figure \ref{fig:GDAUG_Ex}) required for supervised training, such as with the Multi-Task Pretraining paradigm.
        This pipeline provides a reproducible, automated system for generating consistent, chip-aligned supervision across global-scale imagery datasets, bridging open vector sources and high-resolution raster training data. 
         
        While \gls{g-daug} scales the raw quantity of supervision possible, these weak labels introduce domain-specific risks, such as \gls{osm} sparsity in rural regions, temporal lag relative to imagery acquisition, and geometry inconsistencies.
        We mitigate these through ontology curation and resolution-aware rasterization, label noise is a real factor at scale and is reflected in our emphasis on conservative \gls{lr} schedules and checkpointing.

\section{Experimental Setup}
    We run several controlled experiments aligned to our guiding questions: (a) critical batch size sweeps for early- to mid-stage training; (b) \gls{lr} sweeps under a \gls{wsd} schedule with stabilized checkpointing; (c) data scaling tests using nested, stratified subsets from 5k to 1M images; and (d) parameter scaling tests, from 86 million parameters to 1.8 billion parameters. 
    We hold architecture and training dynamics constant within each experiment to isolate the lever under study, and we report both positive trends and negative outcomes that informed our findings.
   \subsection{Model Architectures}
        We leverage two different model configurations for our experiments: \gls{mtp}, which is fully supervised and uses multi-task optimization across multiple label types; and \gls{scalemae}, which is based on self-supervision and scale invariance.
   
        \gls{mtp} is a training paradigm that consists of a single backbone (or shared encoder) and a set of task-specific decoders \cite{mtp}. 
        During pretraining, the model is optimized jointly on three remote sensing tasks: instance segmentation, semantic segmentation, and rotated object detection, all of which are supervised with mask and bounding box labels, as appropriate for each task.
        The shared encoder produces a multi-scale feature pyramid that serves as input to each of the task-specific decoders.
        This framework is targeted towards large foundation models, supporting both \gls{cnn} and \gls{vit} architectures. 
        Additionally, when a \gls{vit} is used, the traditional multi-head full attention is replaced by \gls{rvsa} which helps to account for the various orientations of \gls{rs} objects by rotating and rescaling local windows, allowing the transformer to attend to objects at arbitrary orientations and scales. 
        After pretraining is completed, the model can be finetuned on a downstream \gls{rs} task. 
        
        \gls{scalemae} is a self-supervised pretraining framework designed for multiscale geospatial imagery \cite{scalemae}.
        It extends the standard \gls{mae} by encoding real-world scale information, such as \gls{gsd}, directly into the positional embeddings to enable consistent representations across sensors and resolutions.
        The architecture uses a \gls{vit} encoder and lightweight decoder trained to reconstruct masked image patches, enforcing scale-invariant feature learning through reconstruction losses measured relative to physical, rather than pixel, scale.
        This approach allows the model to learn semantically aligned features across heterogenous imagery without requiring resampling or manual normalization.
        Once pretrained, the encoder can be finetuned for downstream \gls{rs} tasks. 
        
    \subsection{Balancing Training Time and Compute Efficiency}
        To evaluate the relationship between batch size, training time, and compute efficiency, we conducted a series of critical batch size experiments using \gls{scalemae} \gls{vit}-Base architecture \cite{scalemae}.
        This smaller model configuration was chosen to enable multiple experimental repetitions across varying batch sizes and random seeds while maintaining computational feasibility. 
        Each run used the same dataset, hyperparameters, and optimizer (AdamW), ensuring consistent optimization setting across runs.
        During these runs training was performed on a single NVIDIA DGX node containing eight H100 \glspl{gpu} connected by NVLink, progressing through early to mid-stage training before hitting a target validation loss of 0.07 \gls{mse} to capture the regime where batch scaling effects are most prominent.
        Batch sizes selected were 128, 256, 384, 512 and results were averaged over multiple seeds to mitigate variance (see figure \ref{fig:critical_batch}). 
        These settings isolate our ability to manipulate training throughput without confounding optimizer dynamics, enabling a direct read on whether batch scaling accelerates convergence. We report the outcomes and implications for compute budgeting in Section 5.1.
    
    \subsection{Optimal \gls{lr} Scheduling} 
\begin{figure}[!h]
            \centering
            \includegraphics[width=\linewidth]{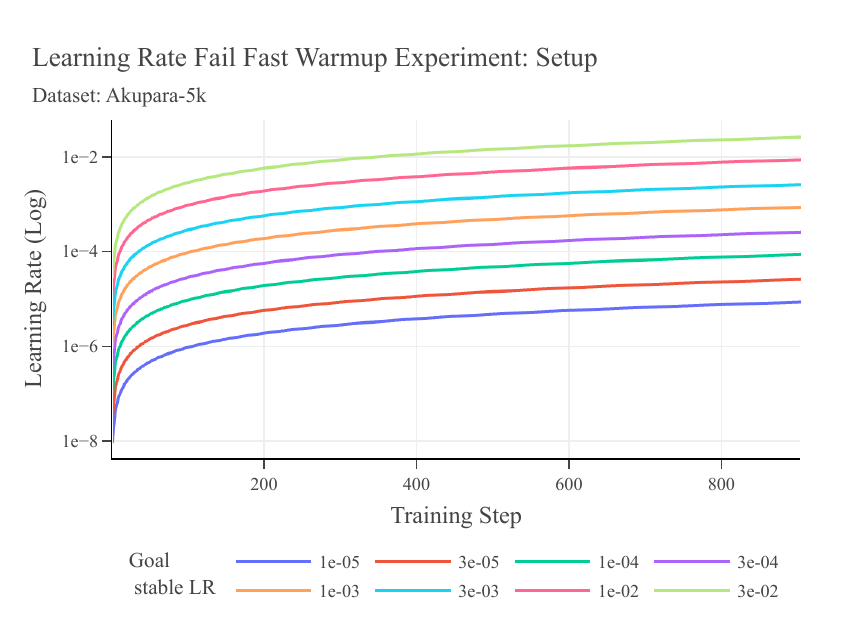}
            \caption{We test eight candidate base learning rates in a fail fast warmup on Akupara-5k, under identical data, batch size, and optimizer settings on four DGX nodes. Each trajectory ramps toward its labeled goal stable learning rate; candidates that exhibit loss spikes, oscillation, or rising gradient norms in this horizon are rejected, while those showing smooth loss decrease and bounded gradient statistics advance to full \gls{wsd} training.}
            \label{fig:fail_fast}
        \end{figure}

\begin{figure}[!h]
            \centering
            \includegraphics[width=\linewidth]{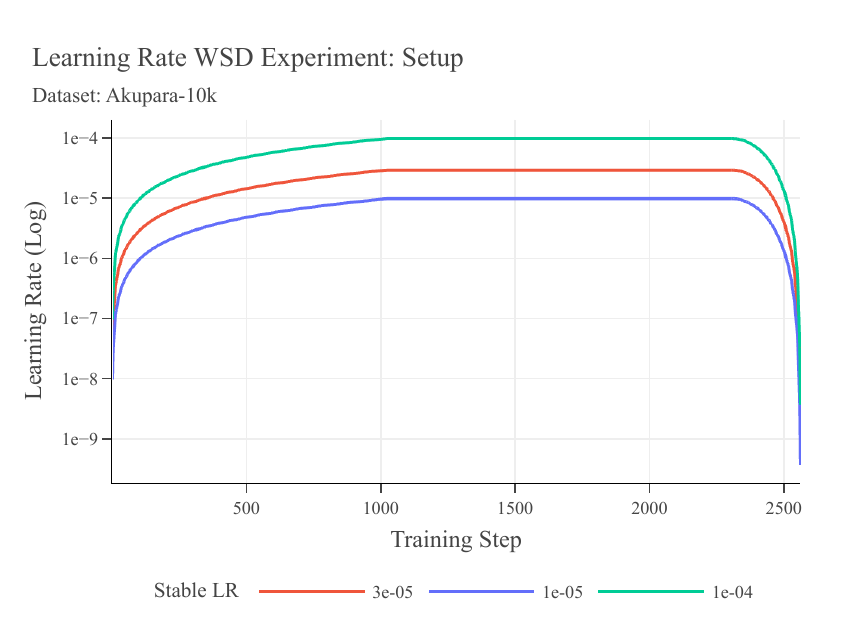}
            \caption{We train three selected stable learning rates with the \gls{wsd} schedule on Akupara-10k, comprising a 10\% warmup, an 80\% stable plateau at the target rate, and an exponential decay initiated when approximately 10\% of the data remains. The curves trace the full \gls{wsd} cycle for each rate, illustrating the transition from warmup to plateau (listed stable \gls{lr}) and subsequent decay.}
            \label{fig:wsd}
        \end{figure}

        To determine ranges of functional learning rates for large-scale geospatial pretraining, we conducted a controlled sweep across multiple rates for \gls{mtp}-trained models.
        Models were trained with the same dataset, batch size and \gls{wsd} scheduling strategy with stabilized checkpointing to isolate the effect of the base learning rate \cite{wsd}.
        Following best practices for responsible compute allocation, we adopt a fail fast protocol to triage candidate base learning rates before committing full \gls{wsd} runs.
        Concretely, we evaluate a broader set of \gls{lr} candidates through 1.6k warmup iterations with Akupara-5k, under identical data, batch, and optimizer settings on four DGX  nodes (32 H100s). The rates used during this preliminary test are shown in figure \ref{fig:fail_fast}.
        Rates that exhibit loss spikes, oscillation, or rising gradient norms in this short horizon are discarded; those that maintain smooth loss decrease and bounded gradient statistics graduate to full \gls{wsd} training with Akupara-10k. 
        The \gls{wsd} schedule maintained an initial warmup period 10\%, followed by a stable plateau phase 80\%, and finally exponential decay initiated when roughly 10\% of the data remained, as shown in figure \ref{fig:wsd}.
        This protocol surfaces instability early, avoids wasting HPC time on doomed configurations, and concentrates full-scale trials on stable candidates.
        The results of both our fail fast phase and our refined \gls{wsd} runs appear in section \ref{section:lr_results}.

    \subsection{Establishing Power Laws for Data Constrained Training}
        To investigate how dataset scale influences model performance in geospatial foundation models, we constructed a series of progressively larger datasets sampled from the Akupara-1M dataset ranging from 5k to 100k full sized satellite images.
        As discussed in section \ref{section:data}, each dataset was sampled to maintain an even distribution across key attributes (location, sensor type, time of day, month, year, land cover class, sensor version, and year), ensuring consistent representation across scales.
        To minimize content variance between sets each successive dataset supersets the smaller ones (i.e., the 10k dataset contains all samples within the 5k dataset). 

        For all experiments we trained \gls{scalemae} \gls{vit}-Giant (\gls{vit}-G, 1.8B parameters) under identical hyperparameters, batch size, and optimization settings.
        This large backbone size was selected to ensure that learning was not limited by model capacity, only by dataset size.
        Training was conducted on a single H100 node (8 GPUs) for approximately 300k iterations per run.
        This controlled setup isolates the effect of dataset size while holding model capacity, compute and training dynamics constant.
        The resulting validation loss curves over training time are shown in figure \ref{fig:dataconstrainedperformance} in the results section.

        This design aims to test whether \gls{rs} follows a stable data-scaling law within practical ranges and to quantify the exponent relevant for planning.
        Specifically, let $N$ denote the number of full-scene images and $L(N)$ denote the converged validation loss under a fixed architecture and schedule. We test the canonical power-law model $L(N) = A + B N^{-a}$, where $A$ is an asymptotic loss floor, $B$ is a scale factor, and $a$ is the data-scaling exponent.
        To obtain $a$, we perform log-log regression on $log(L)$ as a linear function of $log(N)$; we interpret slopes, variance, and residuals from this fit in section \ref{section:power_data_results} and revisit their implications in section \ref{section:discussion}.
        
    \subsection{Establishing Power Laws for Parameter Constrained Training}
\begin{figure}[h!]
            \centering
            \includegraphics[width=1.0\linewidth]{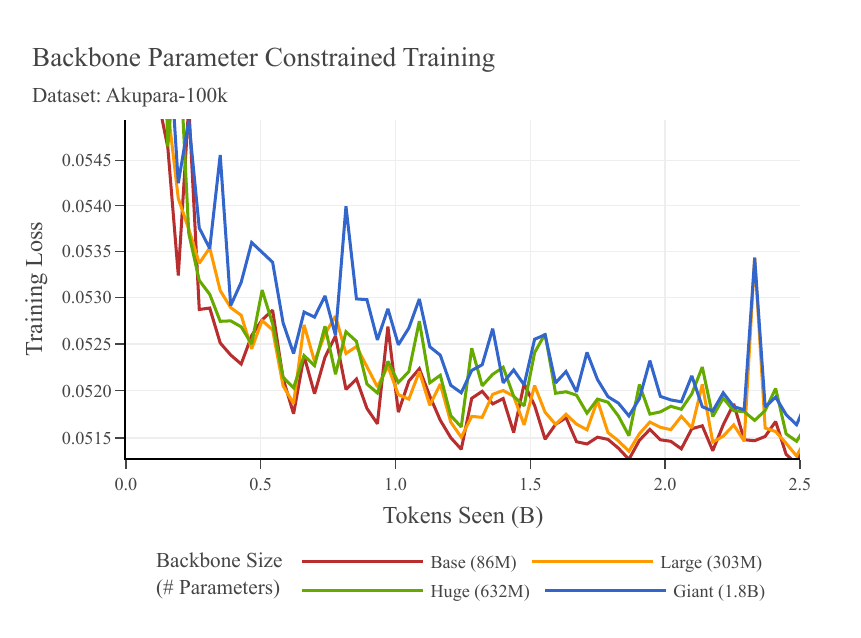}
            \caption{Training curves for identical data, optimizer, and schedule plotted as validation loss versus tokens seen. Despite a range from 86M-1.8B parameters, curves are nearly indistinguishable. This indicates capacity parity under a fixed regime and increasing parameters does not yield lower loss when the dataset and training budget are held constant.}
            \label{fig:backboneconstrainedperformance}
        \end{figure}

        To quantify how encoder capacity alone affects pretraining performance under fixed data and optimization dynamics, we constructed a parameter-constrained experiment that varies \gls{vit} backbone size, spanning increasing parameter counts across orders of magnitude (i.e., \gls{vit}-Base 86M \cite{crossscale}, \gls{vit}-Large 303M \cite{scalemae}, \gls{vit}-Huge 630M \cite{scalingvits}, \gls{vit}-Giant 1.8B \cite{scalingvits}).
        To make the comparison with the data-constrained study explicit, we treat parameter scaling as a one-factor ablation: fix dataset composition, training schedule, and optimizer, and vary only backbone capacity.
        Let $P$ denote encoder parameters and let $L(P)$ denote the converged validation loss under a fixed training budget.
        Consistent with prior neural scaling formulations, we test whether $L(P)$ follows a power-law of the form $L(P)=A+B P^{-b}$, where $A$ approximates an asymptotic floor under the chosen data and supervision regime, $B$ is a scale factor, and $b$ indicates benefit from increased capacity.
        Once again, we perform linear regression on $log(L)$ vs. $log(P)$ to obtain a value for $b$.
        Refer to figure \ref{fig:backboneconstrainedperformance}, section \ref{section:power_backbone_results}, and section \ref{section:discussion} for loss curves, scaling analysis, and discussion of this experiment, respectively.
  
\section{Results}
    Each of the following subsections addresses one of our guiding questions, emphasize the practical takeaways, and the conditions under which the observations hold.
    \subsection{Balancing Training Time and Compute Efficiency}
        \label{section:critical_results}
\begin{figure}[!h]
            \centering
            \includegraphics[width=\linewidth]{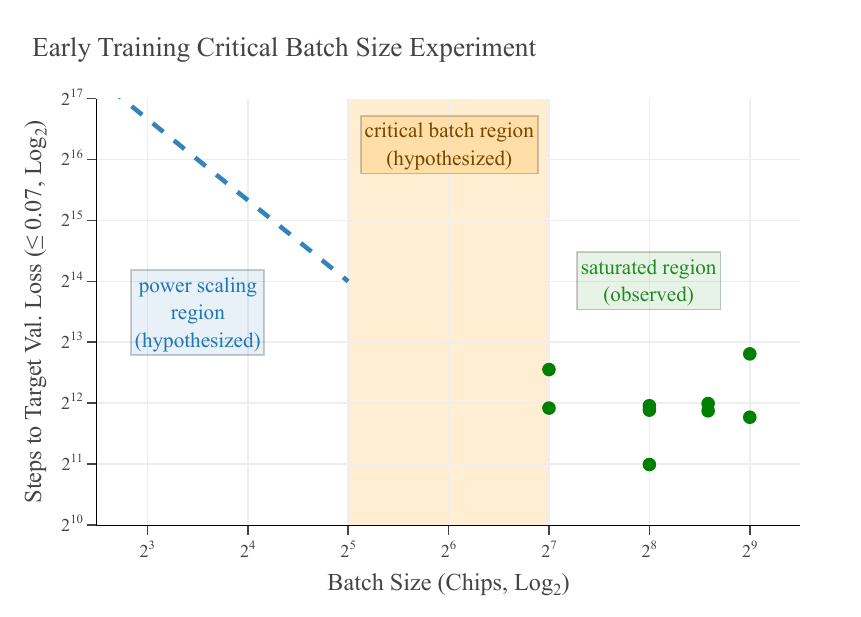}
            \caption{Batch size scaling results show an increase in batch size does not result in increased compute efficiency during early-stage training, based on average loss over three trials. Based on the similar number of steps needed to reach convergence for the \glspl{lr} tested, we hypothesize that smaller batch sizes would maneuver this training configuration out of the saturated range and decrease forward pass compute efficiency.}
            \label{fig:critical_batch}
        \end{figure}

        Across multiple random seeds and batch sizes ranging from 128 to 512, the results shown in figure \ref{fig:critical_batch} indicate little to no improvement in training efficiency or improved convergence when increasing batch size during early- to mid-stage training.
        Variance over 3 trials across seeds remained low, and most models reached the target loss in a similar number of steps, indicating batch size had minimal influence on convergence dynamics in these cases.
        Models trained with larger batches exhibited similar, or occasionally slower, convergence compared to those with moderate batch sizes, suggesting that the critical batch size beyond which further increases yield diminishing returns in sample efficiency falls within or below this tested range.
        These findings are consistent with observations from language and vision foundation models, where critical batch size tends to depend more strongly on dataset scale than on model size \cite{nlpscaling} \cite{revistneuralscaling}.
        While larger batches may still offer benefits in late-stage training, when optimization stabilizes and gradient noise decreases, the current results suggest that early-stage training in geospatial foundation models remains batch-insensitive beyond moderate thresholds.
        This behavior reinforces that compute efficiency in large-scale pretraining should be sought primarily through data scaling and adaptive learning-rate scheduling, rather than indiscriminate batch size increases, which may improve throughput but introduce inefficiencies due to reduced gradient noise. 
        
        \textbf{Takeaway}: In \gls{rs} pretraining, early- to mid-stage convergence is largely insensitive to batch increases beyond moderate sizes; seek efficiency via data diversity and \gls{lr} scheduling rather than aggressive batching.
        
    \subsection{Optimal Learning Rate (LR)}
        \label{section:lr_results}
\begin{figure}[!h]
            \centering
            \includegraphics[width=\linewidth]{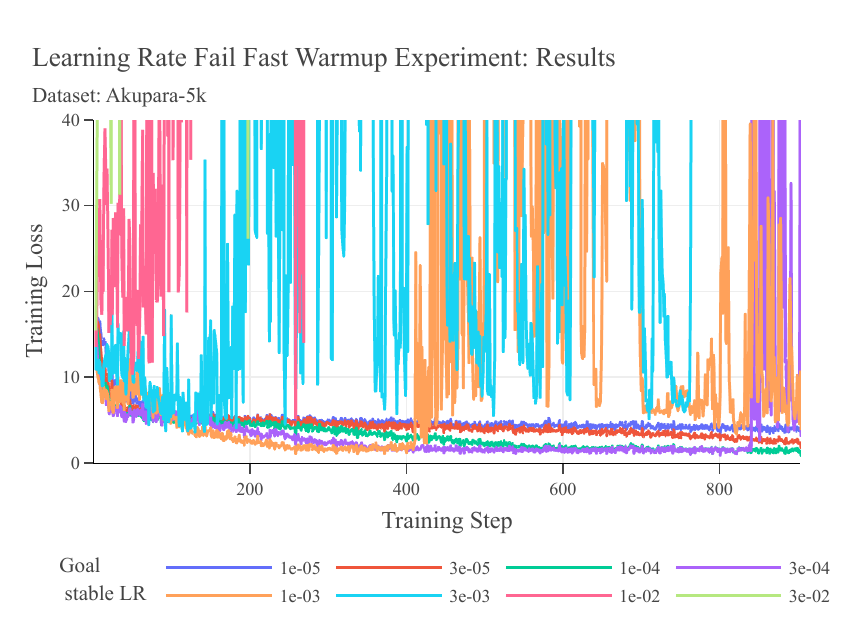}
            \caption{We triage eight learning rate candidates via a fail fast warmup experiment monitoring training loss on Akupara-5k with an \gls{mtp} architecture. Higher rates \(1\times10^{-3}\) and \(3\times10^{-3}\) exhibit significant instability, with divergent or oscillatory loss, the intermediate \(3\times10^{-4}\) achieves acceptable but slower convergence, and moderate candidates—most notably \(1\times10^{-4}\)—show smooth, monotonic loss reduction.}
            \label{fig:results_fast}
        \end{figure}

        Our fail fast preliminary warmup test demonstrated that higher learning rates (e.g., \(1\times10^{-3}\) and \(3\times10^{-3}\)) exhibited significant instability, often diverging or oscillating during the stable phase of training, as shown in figure \ref{fig:results_fast}. 
        Intermediate rates achieved stable convergence, confirming classic intuition that aggressive step sizes reduce optimization stability, and identifying a safe range that allows us to minimize the risk of catastrophic divergence without unnecessarily slowing training.
        
\begin{figure}[!h]
            \centering
            \includegraphics[width=\linewidth]{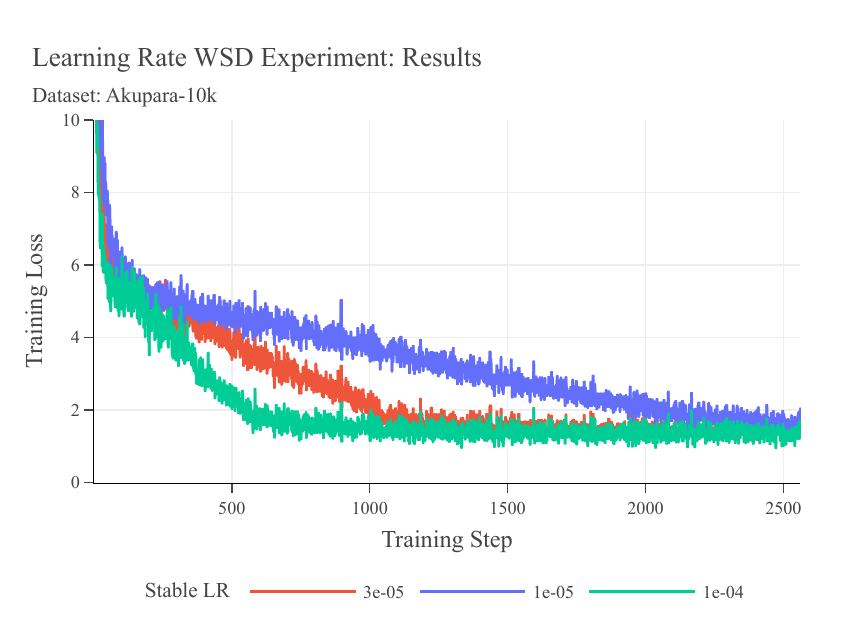}
            \caption{Three learning rate candidates from our fail fast warmup experiment progress to a test of the full \gls{wsd} schedule on Akupara-10k. Among the tested rates, \(1\times10^{-4}\) yields the most stable and efficient progress, maintaining monotonic loss decrease through the plateau and achieving the best final loss after decay, confirming that moderate base rates under \gls{wsd} provide robust optimization, validate the predictive utility of the fail fast selection, and establish a reliable baseline for large-scale \gls{rs} pretraining.}
            \label{fig:results_wsd}
        \end{figure}

        Based on the results of our fail fast test, we graduated three learning rates to a full \gls{wsd} cycle experiment; these results identify \(1\times10^{-4}\) as a functional learning rate for \gls{mtp} with a \gls{vit}-L backbone while under \gls{wsd} scheduling, as shown in figure \ref{fig:results_wsd}.
        The finding aligns with prior scaling studies in language and vision where stable, moderate learning rates yield the best efficiency-performance balance across dataset sizes \cite{revistneuralscaling}.
        
        In practice, this provides a reliable baseline for future large-scale \gls{rs} pretraining.
        Operationally, the fail fast triage eliminated unstable high-\gls{lr} settings within 2,000 steps, saving multi-node hours that would otherwise be consumed by oscillatory runs.
        The \gls{lr} that ultimately performed best at full scale first demonstrated stable, monotonic loss reduction in the fail fast phase, validating the protocol’s predictive utility.
        
        \textbf{Takeaway}: A moderate base \gls{lr} under \gls{wsd} yields robust progress; fail fast warmup sweeps quickly rule out unstable rates and save scarce HPC time, making conservative schedules plus early triage the safer default at \gls{rs} scale.

    \subsection{Establishing Power Laws for Data Constrained Training}
        \label{section:power_data_results}
\begin{figure}[h!]
            \centering
            \includegraphics[width=\linewidth]{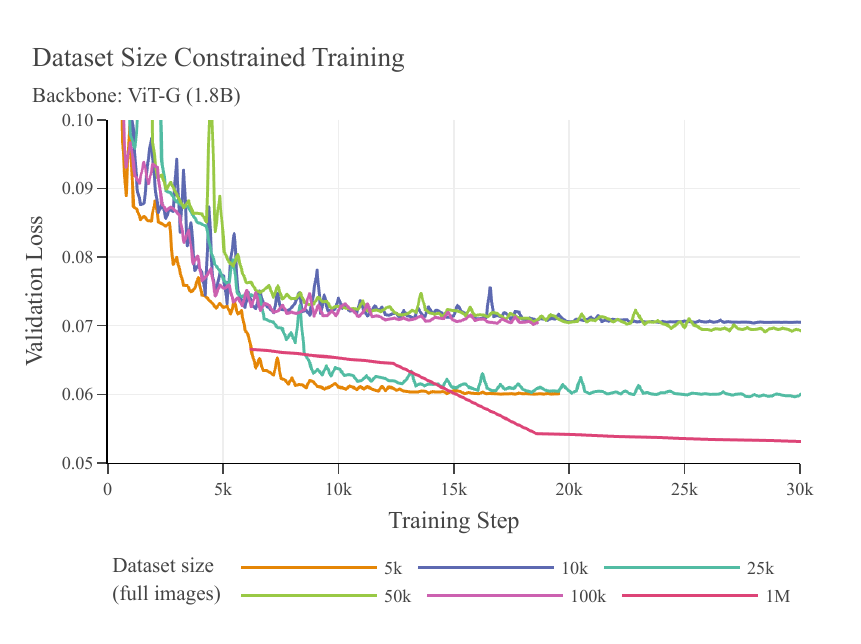}
            \caption{We measure validation loss across the first 30k training steps for six dataset sizes of full satellite images, each trained with an intentionally oversized \gls{vit}-G backbone to ensure data-constrained rather than parameter-constrained behavior. Approximately half of the runs become trapped in a local minimum after the first 5k steps, whereas the remaining runs continue to improve beyond this point.}
            \label{fig:dataconstrainedperformance}
        \end{figure}

\begin{figure}[h!]
            \centering
            \includegraphics[width=\linewidth]{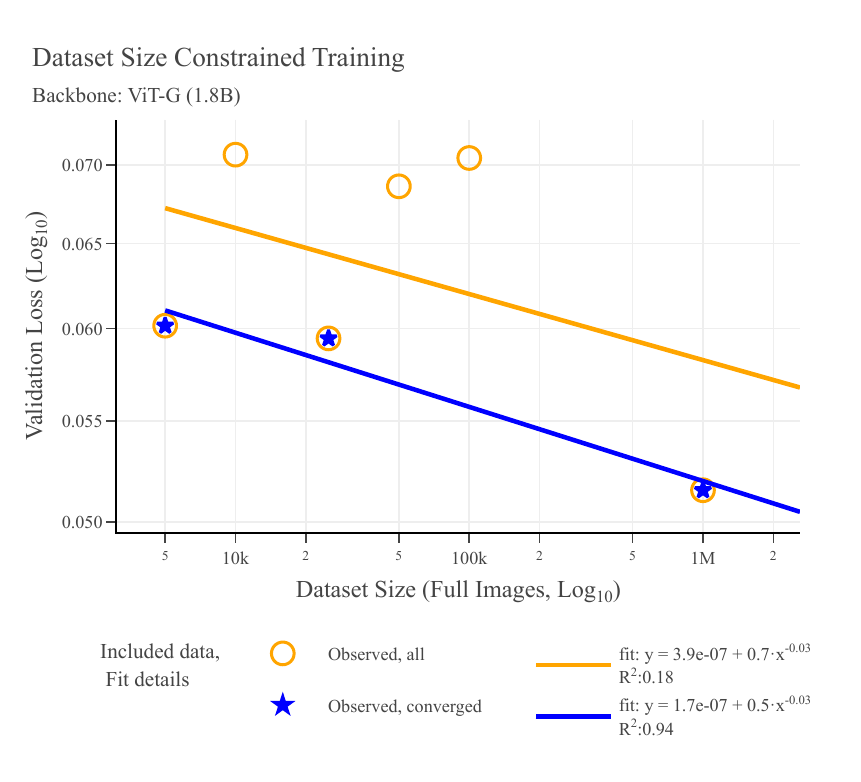}
            \caption{We observe a log–log relationship between eventual validation loss and dataset size under a data-constrained regime. A power-law fit across all six dataset sizes shows poor correspondence ($R^2=0.18$), driven by trials that remain stuck in the early local minimum; restricting the analysis to the three trials that surpassed the minimum yields a strong power-law fit ($R^2=0.94$), enabling performance extrapolation from small-scale experiments to larger datasets.}
            \label{fig:data_scaling_law}
        \end{figure}

        As shown in figure \ref{fig:dataconstrainedperformance}, models exhibited two types of convergence behavior: one subset exhibited anomalously high and plateaued loss values inconsistent with the smooth declining loss trend of remaining experiments.
        Because such optimization failures obscure the underlying scaling behavior, we exclude these points from our fitted curve in figure \ref{fig:data_scaling_law} and plan to rerun these configurations with alternative random seeds to determine whether their performance aligns with the established scaling law. 
        
        When plotting the final converged loss against dataset size on a log-log scale in figure \ref{fig:data_scaling_law}, the results followed a near-linear trend consistent with a power-law relationship between loss and data scale, $L(N)=A+B N^{-a}$.
        The fitted log(loss) scaling exponent was $a=0.03$, with $\mathrm{R}^2=0.941$ after excluding trapped trials, indicating diminishing yet consistent returns with increasing $N$.
        For planning, this analysis provides a closed-form estimator of required scale: given a target loss $L^*$ above baseline $A$, the implied dataset size is $N\approx \left( \frac{L^*-A}{B} \right)^{-1/a}$.
        Although the current fit is based on three scales ranging from 5k - 1M samples, it already demonstrates the hallmark of empirical scaling behavior: consistent slope, low variance, and strong log-log linearity.
        Including additional intermediate points will further test whether this power-law trend persists across multiple orders of magnitude.
        If the slope remains stable as new points are added, even with a modest reduction in $\mathrm{R}^2$, this will strengthen the evidence for a true scaling law rather than a local fit.
        
        \textbf{Takeaway}: Within 5k–1M images, \gls{rs} pretraining follows a stable, diminishing-returns power law, providing a quantitative basis for planning data expansions and estimating expected gains.
        
    \subsection{Establishing Power Laws for Parameter Constrained Training}
        \label{section:power_backbone_results}
\begin{figure}[h!]
            \centering
            \includegraphics[width=\linewidth]{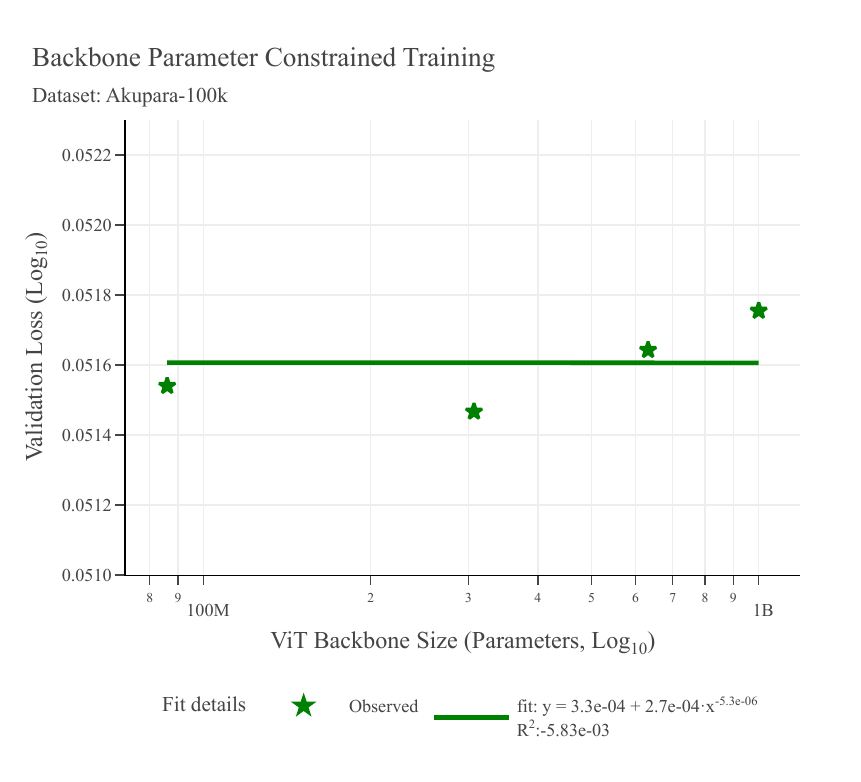}
            \caption{Scaling law fit of converged validation loss against parameter count yields a slope of $b\approx 0$, i.e., no measurable dependence on model size in this regime. The flat trend is consistent with a data-limited regime rather than a model size limited one.}
            \label{fig:model_scaling_law}
        \end{figure}

        Under our fixed-budget setup, the estimated slope is statistically indistinguishable from zero, i.e., $b\approx 0$ (see figure \ref{fig:model_scaling_law}), indicating that capacity increases do not reduce loss when data, schedule, and optimization are held constant.
        This outcome is consistent with a data-limited regime and also reflects common confounders that arises when parameters are scaled without adjusting training budget or context.
        Specifically, with identical steps and batch size, larger models receive fewer training tokens ($T$) per parameter; when scaling, optimal allocation maintains $T/P$ above a domain-dependent threshold and holding $T$ fixed while increasing $P$ typically suppresses parameter-scaling gains.
        Additionally, \gls{vit} patch sizes may need adjustment as backbone capacity scales to expose sufficient context and data structure at training time. 
        Similarly, \gls{lr} schedules derived empirically from training runs of smaller models or based solely on dataset size may result in under-training of larger backbones.
        
        To separate these effects from true parameter-scaling limits, future studies would benefit from compute-aware parameter-scaling protocols to avoid inadvertently penalizing larger models.
        For example, keeping tokens-per-parameter approximately constant across backbones guards against under-training larger models.
        With such controls in place, a parameter-scaling exponent $b>0$ would indicate genuine gains from capacity, whereas $b\approx 0$ would confirm that, even under compute-matched conditions, \gls{rs} pretraining remains data-limited for the given supervision and context.
        
        \textbf{Takeaway}: In our fixed-budget study, the near-flat trend supports prioritizing in-domain data expansion and schedule robustness over parameter scaling. To this end, future studies should expand this investigation by repeating it with larger training datasets to reassess the computed scaling exponent.

\section{Discussion}
    \label{section:discussion}
    Our findings demonstrate that large-scale \gls{rs} foundation models follow predictable scaling dynamics, like those of related domains, and also exhibit domain-specific behaviors.
    The power-law established with an $\mathrm{R}^2$ fit of 0.941 between the dataset size and performance confirms that additional data consistently improves model quality, though the exponent $a=0.03$ indicates diminishing returns as datasets grow, consistent with trends observed in natural imagery and language domains \cite{nlpscaling} \cite{scalingvits}. 
    Efficiency analyses reveal that compute gains from batch-size scaling plateau quickly and that training stability depends more on balanced learning rate schedules than raw throughput.
    These findings imply that further performance gains will come less from scaling compute and more from improving data diversity, label quality, and optimization strategy.
    Overall, \gls{rs} models appear to operate in a data-limited, yet feature-bounded regime.
    Scaling continues to improve training progress, but only as it expands the domain of observed geospatial variation.
    Future scaling should therefore emphasize use of more feature-rich sensing modalities (e.g., \gls{sar}, multispectral, temporal), and geographically or contextually novel data, rather than simple quantity increases.
    The shallow parameter-scaling slope in figure \ref{fig:model_scaling_law} relative to data-scaling in figure \ref{fig:data_scaling_law} supports the view that \gls{rs} pretraining is data-limited: capacity helps, but only insofar as it can absorb diverse, in-domain variation. In practice, prioritize expanding and domain coverage, then right-size to avoid undercapacity in rich regimes and overcapacity in narrow regimes.
    
    \subsection{Lessons learned and pitfalls}
    \begin{itemize}
        \item Batch-size illusions: Larger batches did not accelerate early/mid-stage convergence; late-stage benefits are limited and context-bounded.
        \item \gls{lr}-induced instability: High base \glspl{lr} under \gls{wsd} produced oscillations and divergence; conservative \glspl{wsd} reduced instability.
        \item Bounded spatial context: Increasing model size without expanding image context or modality yields marginal returns; invest in diversity and richer sensing (\gls{sar}, \gls{msi}, temporal).
    \end{itemize}
    
    \subsection{Scope limitations}
        Our results reflect high-resolution imagery under \gls{vit} backbones and specific schedules; different sensors (e.g., \gls{sar}), data distributions, or architectural choices (e.g., \gls{cnn} hybrids) may alter constants even if the qualitative trends persist. 

\section{Conclusion}
    This work establishes empirical scaling behavior for \gls{rs} foundation models trained at peta-scale data volumes.
    We show that model loss decreases predictably with dataset size,while gains from larger batches or aggressive learning rates remain limited.
    These results suggest that sustainable progress in \gls{rs} foundation modeling will hinge on use of more diverse data, not merely more compute. 
    By quantifying how performance scales with data, this study provides a practical basis for planning future pretraining efforts and estimating the returns on additional data collection.
    Extending this framework to multimodal, multi-temporal, and cross-sensor contexts offers a clear next step toward globally consistent geospatial representation learning.
    Future studies should compare training performance, as measured by optimization task loss, to downstream \gls{rs} task performance to characterize the generalizability of our demonstrated scaling laws.
    \vspace{10pt}
    
    \noindent \textbf{Practical checklist for \gls{rs} scaling}:
    \begin{itemize}
        \item Start with stability: choose conservative \glspl{lr} under \gls{wsd}; enable stabilized checkpointing.
        \item Right-size batches: avoid aggressive early-stage batch scaling; monitor convergence per token, not just throughput.
        \item Expand diversity, not just volume: stratify by geography, sensor, season, and context to broaden feature coverage.
        \item Anticipate label noise: curate ontologies, align \gls{gsd}, and validate regional coverage; expect residual noise and plan schedules accordingly.
        \item Measure scaling pragmatically: fit simple power laws over controlled ranges and use them to plan data collection and compute budgets.
        \item Fail fast on hyperparameter sweeps: triage a broad \gls{lr} set through warmup iterations; promote only stable candidates to full \gls{wsd} runs to avoid burning multi-node allocations.
\end{itemize}

\begin{credits}
    \subsubsection{\ackname}
        This work was supported by the MITRE Independent Research and Development Program.

    \subsubsection{\discintname}
        The authors have no competing interests to declare that are relevant to the content of this article.

    \subsubsection{NOTICE.}
        Portions of this technical data were produced for the U. S. Government under Contract No. FA8702-19-C-0001 and W56KGU-18-D-0004, and is subject to the Rights in Technical Data-Noncommercial Items Clause DFARS 252.227-7013 (FEB 2014)\\
\end{credits}
\begin{center}
    \textbf{\textcopyright 2025 The MITRE Corporation}
\end{center}

\raggedright \textbf{Approved for Public Release; Distribution Unlimited. Public Release Case Number 25-2674.}
    
\bibliographystyle{splncs04}
\bibliography{scaling}

@inproceedings{scalemae,
    author = {Reed, Colorado and Gupta, Ritwik and Li, Shufan and Brockman, Sarah and Funk, Christopher and Clipp, Brian and Keutzer, Kurt and Candido, Salvatore and Uyttendaele, Matt and Darrell, Trevor},
    year = {2023},
    month = {10},
    pages = {4065-4076},
    title = {{Scale-MAE}: A Scale-Aware Masked Autoencoder for Multiscale Geospatial Representation Learning},
    urldate={2023-10-01},
    url="https://doi.org/10.1109/ICCV51070.2023.00378"
}

@ARTICLE{mtp,
  author={Wang, Di and Zhang, Jing and Xu, Minqiang and Liu, Lin and Wang, Dongsheng and Gao, Erzhong and Han, Chengxi and Guo, Haonan and Du, Bo and Tao, Dacheng and Zhang, Liangpei},
  journal={{IEEE} Journal of Selected Topics in Applied Earth Observations and Remote Sensing}, 
  title={{MTP}: Advancing Remote Sensing Foundation Model via Multitask Pretraining}, 
  year={2024},
  volume={17},
  number={},
  pages={11632-11654},
  url="https://doi.org/10.1109/JSTARS.2024.3408154",
  urldate={2024-01-01}
}

@misc{satlaspretrain,
    title={{SatlasPretrain}: A Large-Scale Dataset for Remote Sensing Image Understanding}, 
    author={Favyen Bastani and Piper Wolters and Ritwik Gupta and Joe Ferdinando and Aniruddha Kembhavi},
    year={2023},
    eprint={2211.15660},
    archivePrefix={arXiv},
    primaryClass={cs.CV},
    url="https://arxiv.org/abs/2211.15660", 
    urldate={2023-01-01}

}

@misc{prithvi,
    title={Prithvi-{EO}-2.0: A Versatile Multi-Temporal Foundation Model for Earth Observation Applications}, 
    author={Daniela Szwarcman and Sujit Roy and Paolo Fraccaro and Þorsteinn Elí Gíslason and Benedikt Blumenstiel and Rinki Ghosal and Pedro Henrique de Oliveira and Joao Lucas de Sousa Almeida and Rocco Sedona and Yanghui Kang and Srija Chakraborty and Sizhe Wang and Carlos Gomes and Ankur Kumar and Myscon Truong and Denys Godwin and Hyunho Lee and Chia-Yu Hsu and Ata Akbari Asanjan and Besart Mujeci and Disha Shidham and Trevor Keenan and Paulo Arevalo and Wenwen Li and Hamed Alemohammad and Pontus Olofsson and Christopher Hain and Robert Kennedy and Bianca Zadrozny and David Bell and Gabriele Cavallaro and Campbell Watson and Manil Maskey and Rahul Ramachandran and Juan Bernabe Moreno},
    year={2025},
    eprint={2412.02732},
    archivePrefix={arXiv},
    primaryClass={cs.CV},
    url="https://arxiv.org/abs/2412.02732", 
    urldate={2025-01-01}
}

@InProceedings{scalingvits,
    author    = {Zhai, Xiaohua and Kolesnikov, Alexander and Houlsby, Neil and Beyer, Lucas},
    title     = {Scaling Vision Transformers},
    booktitle = {Proceedings of the {IEEE/CVF} Conference on Computer Vision and Pattern Recognition {(CVPR)}},
    month     = {June},
    year      = {2022},
    pages     = {12104-12113},
    urldate={2022-06-01},
    url="https://arxiv.org/abs/2106.04560"
}

@inproceedings{revistneuralscaling,
    author = {Alabdulmohsin, Ibrahim M and Neyshabur, Behnam and Zhai, Xiaohua},
    booktitle = {Advances in Neural Information Processing Systems},
    editor = {S. Koyejo and S. Mohamed and A. Agarwal and D. Belgrave and K. Cho and A. Oh},
    pages = {22300--22312},
    publisher = {Curran Associates, Inc.},
    title = {Revisiting Neural Scaling Laws in Language and Vision},
    url = "https://proceedings.neurips.cc/paper_files/paper/2022/file/8c22e5e918198702765ecff4b20d0a90-Paper-Conference.pdf",
    volume = {35},
    year = {2022},
    urldate={2022-01-01}

}

@ARTICLE{billionscale,
  author={Cha, Keumgang and Seo, Junghoon and Lee, Taekyung},
  journal={{IEEE} Journal of Selected Topics in Applied Earth Observations and Remote Sensing}, 
  title={A Billion-scale Foundation Model for Remote Sensing Images}, 
  year={2024},
  volume={},
  number={},
  pages={1-17},
  url="https://doi.org/10.1109/JSTARS.2024.3401772",
  urldate={2024-01-01}
}

@InProceedings{skysense,
    author    = {Guo, Xin and Lao, Jiangwei and Dang, Bo and Zhang, Yingying and Yu, Lei and Ru, Lixiang and Zhong, Liheng and Huang, Ziyuan and Wu, Kang and Hu, Dingxiang and He, Huimei and Wang, Jian and Chen, Jingdong and Yang, Ming and Zhang, Yongjun and Li, Yansheng},
    title     = {{SkySense}: A Multi-Modal Remote Sensing Foundation Model Towards Universal Interpretation for Earth Observation Imagery},
    booktitle = {Proceedings of the {IEEE/CVF} Conference on Computer Vision and Pattern Recognition {(CVPR)}},
    month     = {June},
    year      = {2024},
    pages     = {27672-27683},
    urldate={2024-06-01},
    url="https://arxiv.org/abs/2312.10115"
}

@misc{dinov3,
    title={{DINOv3}}, 
    author={Oriane Siméoni and Huy V. Vo and Maximilian Seitzer and Federico Baldassarre and Maxime Oquab and Cijo Jose and Vasil Khalidov and Marc Szafraniec and Seungeun Yi and Michaël Ramamonjisoa and Francisco Massa and Daniel Haziza and Luca Wehrstedt and Jianyuan Wang and Timothée Darcet and Théo Moutakanni and Leonel Sentana and Claire Roberts and Andrea Vedaldi and Jamie Tolan and John Brandt and Camille Couprie and Julien Mairal and Hervé Jégou and Patrick Labatut and Piotr Bojanowski},
    year={2025},
    eprint={2508.10104},
    archivePrefix={arXiv},
    primaryClass={cs.CV},
    url="https://arxiv.org/abs/2508.10104", 
    urldate={2025-01-01}
}

@misc{nlpscaling,
    title={Scaling Laws for Neural Language Models}, 
    author={Jared Kaplan and Sam McCandlish and Tom Henighan and Tom B. Brown and Benjamin Chess and Rewon Child and Scott Gray and Alec Radford and Jeffrey Wu and Dario Amodei},
    year={2020},
    eprint={2001.08361},
    archivePrefix={arXiv},
    primaryClass={cs.LG},
    url="https://arxiv.org/abs/2001.08361", 
    urldate={2020-01-01}
}

@inproceedings{attention,
    author = {Vaswani, Ashish and Shazeer, Noam and Parmar, Niki and Uszkoreit, Jakob and Jones, Llion and Gomez, Aidan N and Kaiser, \L ukasz and Polosukhin, Illia},
    booktitle = {Advances in Neural Information Processing Systems},
    editor = {I. Guyon and U. Von Luxburg and S. Bengio and H. Wallach and R. Fergus and S. Vishwanathan and R. Garnett},
    pages = {},
    publisher = {Curran Associates, Inc.},
    title = {Attention is All you Need},
    url = "https://arxiv.org/abs/1706.03762",
    volume = {30},
    year = {2017},
    urldate={2017-01-01}
}

@inproceedings{vits,
    title={An Image is Worth 16x16 Words: Transformers for Image Recognition at Scale},
    author={Alexey Dosovitskiy and Lucas Beyer and Alexander Kolesnikov and Dirk Weissenborn and Xiaohua Zhai and Thomas Unterthiner and Mostafa Dehghani and Matthias Minderer and Georg Heigold and Sylvain Gelly and Jakob Uszkoreit and Neil Houlsby},
    booktitle={International Conference on Learning Representations},
    year={2021},
    url="https://openreview.net/forum?id=YicbFdNTTy", 
    urldate={2021-07-01}
}

@misc{wsd,
    title={Understanding {Warmup-Stable-Decay} Learning Rates: A River Valley Loss Landscape Perspective}, 
    author={Kaiyue Wen and Zhiyuan Li and Jason Wang and David Hall and Percy Liang and Tengyu Ma},
    year={2024},
    eprint={2410.05192},
    archivePrefix={arXiv},
    primaryClass={cs.LG},
    url="https://arxiv.org/abs/2410.05192", 
    urldate={2025-01-01}
}

@InProceedings{mae,
    author    = {He, Kaiming and Chen, Xinlei and Xie, Saining and Li, Yanghao and Doll\'ar, Piotr and Girshick, Ross},
    title     = {Masked Autoencoders Are Scalable Vision Learners},
    booktitle = {Proceedings of the {IEEE/CVF} Conference on Computer Vision and Pattern Recognition {(CVPR)}},
    month     = {June},
    year      = {2022},
    pages     = {16000-16009},
    urldate={2022-06-01},
    url="https://arxiv.org/abs/2111.06377"
}

@inproceedings{crossscale,
    author = {Tang, Maofeng and Cozma, Andrei and Georgiou, Konstantinos and Qi, Hairong},
    booktitle = {Advances in Neural Information Processing Systems},
    editor = {A. Oh and T. Naumann and A. Globerson and K. Saenko and M. Hardt and S. Levine},
    pages = {20054--20066},
    publisher = {Curran Associates, Inc.},
    title = {Cross-Scale {MAE}: A Tale of Multiscale Exploitation in Remote Sensing},
    url = "https://proceedings.neurips.cc/paper_files/paper/2023/file/3fadcbd0437f4717723ff3f6f7216800-Paper-Conference.pdf",
    volume = {36},
    year = {2023}, 
    urldate={2023-11-19}
}

\end{document}